# MultiTask-CenterNet (MCN): Efficient and Diverse Multitask Learning using an Anchor Free Approach

Falk Heuer, Sven Mantowsky, Syed Saqib Bukhari, Georg Schneider

ZF Friedrichshafen AG, Artificial Intelligence Lab, Saarbrücken, Germany

{falk.heuer, sven.mantowsky, saqib.bukhari, georg.schneider}@zf.com

## Abstract

*Multitask learning is a common approach in machine learning, which allows to train multiple objectives with a shared architecture. It has been shown that by training multiple tasks together inference time and compute resources can be saved, while the objectives performance remains on a similar or even higher level. However, in perception related multitask networks only closely related tasks can be found, such as object detection, instance and semantic segmentation or depth estimation. Multitask networks with diverse tasks and their effects with respect to efficiency on one another are not well studied.*

*In this paper we augment the CenterNet anchor-free approach for training multiple diverse perception related tasks together, including the task of object detection and semantic segmentation as well as human pose estimation. We refer to this DNN as Multitask-CenterNet (MCN). Additionally, we study different MCN settings for efficiency. The MCN can perform several tasks at once while maintaining, and in some cases even exceeding, the performance values of its corresponding single task networks. More importantly, the MCN architecture decreases inference time and reduces network size when compared to a composition of single task networks.*

## 1. Introduction

Deep Neural networks (DNNs) are used for several nontrivial, strongly nonconvex problems either as a task assistance or a completely self-sufficient and automatic solution. While some tasks are easily well defined, such as image classification, others are more complex and require a sophisticated data processing pipeline. The construction of a self-driving vehicle or the comprehension of an arbitrary voice command are examples of the latter. These complex tasks are hard to train robustly in an end-to-end (single task) manner [21]. Instead, such tasks can be split up into several subtasks that are feasibly tackled by a DNN and can be well described by a loss function. In a later step these abstractions can be merged and postprocessed for either more subtasks or the final objective. In the example of an autonomous car, the environment perception represents a first task separation into subtasks like detection or image segmentation. Later, a scene prediction in a 3D world model with subtasks like road user behavior prediction and scene understanding describe a second separation. It can then be incorporated in the actuator control of the vehicle.

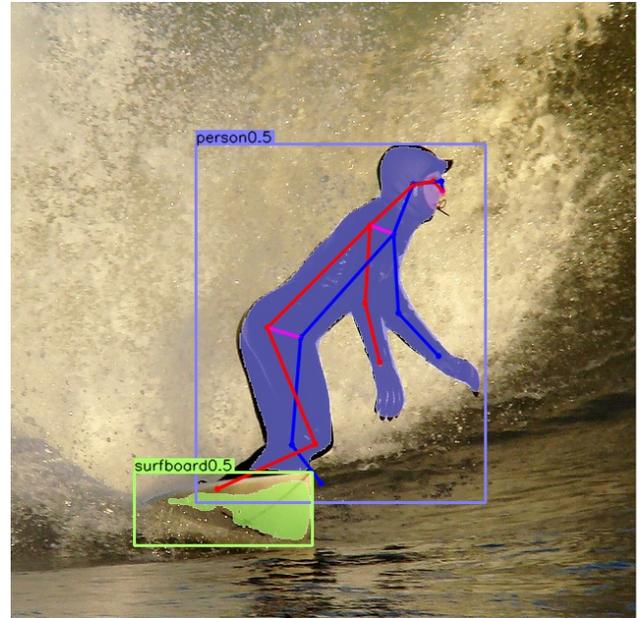

Figure 1: Visualization of a multitask network for object detection, semantic segmentation, and human pose estimation.

Most of the DNN based perception techniques proposed in the literature focus on a single task at a time (like classification, detection, depth estimation, semantic segmentation, pose estimation, etc.). This type of learning mechanism can be referred as Single Task Learning (STL) [7] via Single Task Networks (STN). In contrast to STL, Multitask Learning (MTL) via Multitask Networks (MTN) research [17, 18, 19, 20] has shown that training multiple tasks together in relation to one another not only results is a small DNN, but it can sometimes even enhance the quality of the training and prediction. When networks receive the same kind of input it is likely that similar features will be extracted. A shared backbone can be feasible in such a situation. Additionally, from a hardware perspective, the sharing of feature processing steps can reduce latency and save storage space. In Convolutional Neural Networks (CNNs) for image processing, this can be done by sharing convolutional layers or the entire backbone and split the network at a later network stage into several heads. It is possible to maintain or even increase network performance with such a pipeline by exploiting symbiotic effects [1].

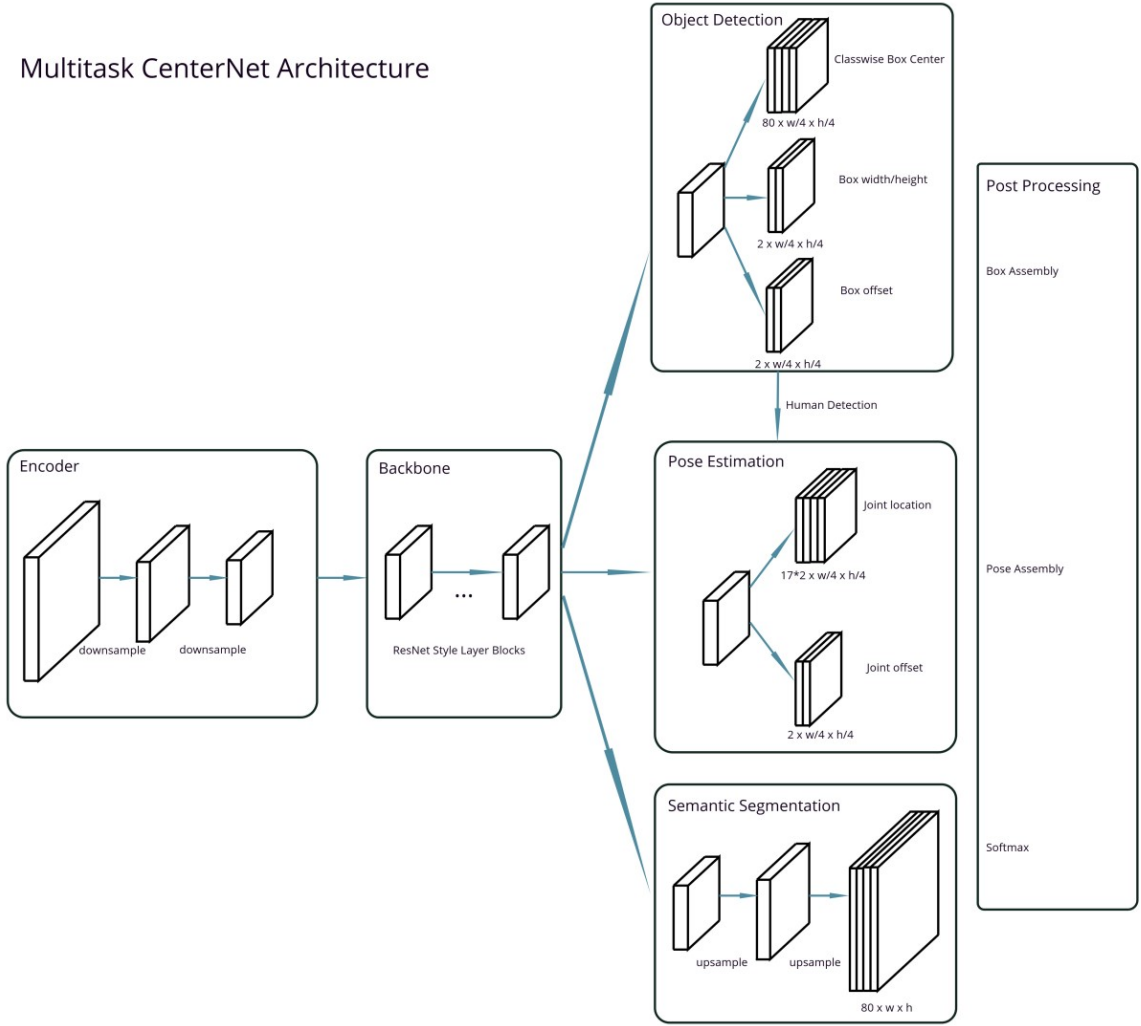

Figure 2: Depiction of the proposed Multitask-CenterNet (MCN). Multiple tasks are performed with a single backbone, saving computation time and resources.

In this paper we augment the anchor-free CenterNet [6] approach to a multitask network called Multitask-CenterNet (MCN) for object detection, semantic segmentation, and human pose estimation together (as depicted in Figure 1). We show that for the MS COCO dataset [22], the proposed multitask network for 2D bounding box detection and semantic segmentation can reach and even outperform mAP values over a singular bounding box detection network. Furthermore, the MCN can be trained on the MS COCO pose estimation task and achieve better pose results when trained together with a semantic segmentation head when compared to a sole pose estimation network. Several other setups of network heads are also studied to explore the coexistence of MTL in a broad diversity. While not all combinations of tasks are equally performant, the saved computation time and network size in the MCN is verifiably efficient.

The rest of the paper is organized as follows. Section 2 highlights some related work for multitask learning from the literature. Section 3 presents the architecture of the proposed Multitask-CenterNet (MCN). Experiments and results are discussed in Section 4. Finally, the paper is concluded in Section 5.

## 2. Related Work

State of the art STL perception networks typically use stacked layers of convolutional filters, represented by a backbone, and a post processing step folding the data into the output domain via the networks head. Some of the widely used techniques for STL for perception tasks are Deeplab [25] for semantic segmentation or YOLO [15] for object detection.

In contrast to that, a multitask network employs several such heads for each task while maintaining a common backbone of layers. The Multitask-CenterNet (MCN) architecture in this work is depicted in Figure 2.

In general, multitask algorithms have been used to merge algorithmic processing in several ways. By keeping a shared representation for several tasks up to late processing within an algorithm, one task can be beneficial for another, as described in [7]. For perception related multitask training there are several different directions. As a flavor of multitask learning, one can consider a pre-trained backbone, which is later used in different pipelines for training different tasks. For example, a DNN backbone is trained on ImageNet [23] and the final layers of the network are exchanged to suit the training pipeline of another task such as object detection. This is beneficial not only for a faster training but also an improved performance when compared to training from random initialization [8]. A closely related direction to multitask learning is where even a single task network consists of multiple losses [6].

Besides the use of multiple heads in MTL, parallel backbone weights with partial or full interconnectivity have been proposed [26]. Heads may be split directly after the backbone or smoothly transition from the backbone by performing an iterative widening of layers [27].

Other occurrences of MTL for perception related problems are Mask RCNN [1], which performs detection and instance segmentation. Panoptic segmentation [12] adds semantic segmentation to instance segmentation and in another work depth estimation have been added to panoptic segmentation [2]. In PersonLab [13] instance segmentation is mutually predicted with human pose estimation. In another multitask network human faces are detected mutually with pose estimation, gender recognition and landmark localization [3].

However, perception related multitask networks usually train a low number of tasks which are often interdependent. In contrast to that, we propose a novel network architecture which jointly trains a large diversity of tasks. Among these are multiclass and single-class object detection, multiclass and single-class semantic segmentation and human pose estimation. The training of constellations of networks with various heads and their influence on performance and resource consumption compared to each other is the focus of our paper.

## 3. The Multitask-CenterNet (MCN)

In this work, multitask networks are studied under the premise of efficiency gains in terms of network size, latency and performance caused by layer sharing as well as their influence on shared training appearing in diverse multitask learning. The MCN architecture is shown in Figure 2. It is explained as follows:

A MCN consists of a backbone of stacked convolutional layers, ReLU activations and batch normalization. It processes the image input. ResNet [14] and DLA [9] backbones are employed. Later processing steps are performed in individual heads after the backbone.

For 2D bounding box detection as well as human pose estimation, CenterNet [6] is used as anchor free detection algorithm in the MCN architecture. Semantic Segmentation is performed with a fully convolutional approach as in FCN [10].

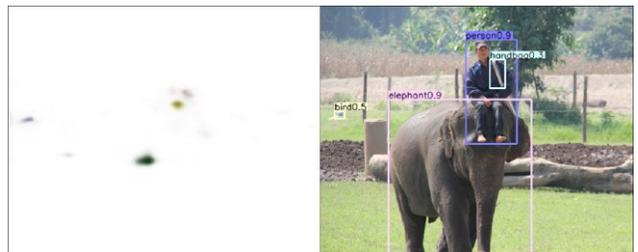

Figure 3: Raw heatmaps from anchor-free detection prediction (left) and their corresponding predicted boxes (right) after box offset, size and thresholding.

Anchor-free detection algorithm like CenterNet can directly use heatmaps generated from feature maps without the necessity for discretization of data with default bounding boxes for bounding box detection. In contrary to that, box detection algorithms such as Mask RCNN [1] or SSD [11] use default bounding boxes followed by non-maximum suppression. In the CenterNet based anchor-free approach, boxes are directly translated into two dimensional gaussian distributions whose maxima mark the boxes center (as shown in Figure 3). Distributions of boxes of one class then form a heatmap as the output of a CNN. The maximum values of predicted heatmaps are used to find box instances. The loss for box detection head can be described with $L_{center}$. Boxes are then refined with another heatmap containing the box pixel size values and regressed to with $L_{size}$. Additionally, the box offset from the heatmaps pixel-grid is calculated with $L_{off}$. For implementation details we refer to [6].

The CenterNet work also presents another similar architecture for human keypoint detection, which is independent of its object detection architecture. In that model, human joints are extracted as keypoints. Keypoints can be detected analogously to box centers with $L_{keyp}$ and their offset is regressed to with $L_{keyp\_off}$.

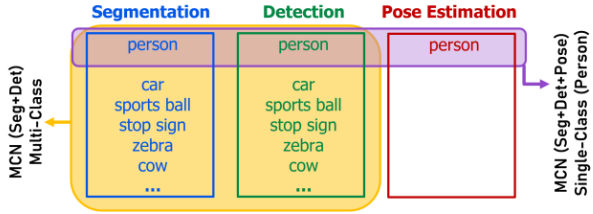

Figure 4: A schematic diagram of the studied output head functionalities. From each task, either one class (person) or, if available, all 80 classes of the MS COCO dataset can be trained in MCN. As an example, multi-class multitask segmentation and detection (without pose estimation) and single-class (only person) multitaks segmentation, detection and pose estimation altogether are shown in green.

In contrary to CenterNet, in the proposed MCN architecture we train a model for object detection for multiple classes together with joint detection for the human class with subsequent heads.

In addition, we also extend our MCN architecture for semantic segmentation together with the other two tasks. In our architecture for semantic segmentation, a pixelwise classification of the image is performed. Feature maps from the backbone are upsampled to the segmentation map size for each predicted class. A softmax layer serves as a normalization over the classes. Its loss can be described with $L_{seg}$ as in [10].

Like the previous multitask mixtures, we train all three of these vision tasks (A, B and C) together in parallel. Anchor-free detection is trained with $L_{center}$, $L_{off}$ and $L_{size}$, while human pose estimation additionally requires $L_{keyp}$ and $L_{keyp\_off}$ besides the box detections for humans. The total loss for the MCN with multiclass detection, multiclass segmentation and human pose estimation is hence described by:

$$L_{MCN} = L_{center} + \lambda_{size} L_{size} + L_{off} + L_{keyp} + L_{keyp\_off} + \lambda_{seg} L_{seg}$$

We choose $\lambda_{size} = 0.1$ and $\lambda_{seg} = 5$ to create an even impact of individual heads on the total loss. Specific performance metrics can benefit or suffer from different weight balancing. Experiments are done with some or all of the tasks and their respective losses. While the balancing has been optimized for this task, we have not performed a grid search over loss balancing and network heads as it would result in an excess of combinatorial computation.

The dataset chosen for all MCNs is the MS COCO dataset [22]. It contains 118k images labeled with instance segmentations of 80 classes, i.e. bounding boxes and segmentation maps per box. To train a network with a semantic segmentation head, instance segmentations of the same class are merged into a common segmentation map. For tasks containing keypoints, the MS COCO keypoint annotations are added. This is only the case for annotations of the class 'person', as keypoints contain human joint data.

Hence, MCNs with one class ('person') or all classes can be trained (Figure 4). A task dissemination is performed to evaluate the influence tasks have on each other, both for different tasks and different class subsets.

## 4. Performance Evaluation and Results

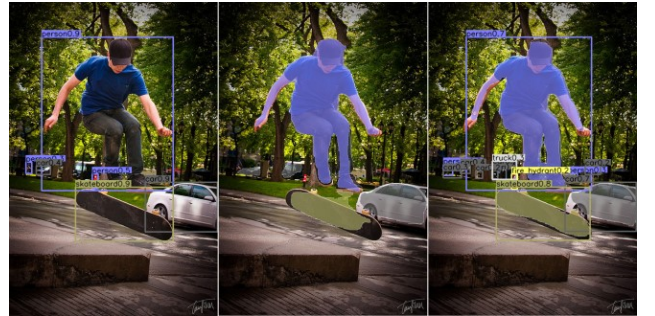

Figure 5: A qualitative analysis of three multiclass MCN. Multiclass detection alone (left), multiclass segmentation alone(center) and multiclass detection + segmentation (right) networks are visualized on a test sample.

Here several MCN network architectures are compared against each other. Unless specifically addressed, the backbone for each MCN is DLA-34 [9]. In a first setup, multi-class detection alone, segmentation alone and detection and segmentation together as multitask are performed. Performance results can be found in Table 1. Notably, during multitask training the performance of segmentation does not decrease significantly even though networks must learn several tasks with the same network backbone. For the detection metric, a multitask network even outperforms a single task network by 0.4% mAP. One sample for the discussed networks can be seen in Figure 5. The single task detection network is identical to CenterNet [6] for which we reproduce the results. The case of single-class triple multitask networks is presented next.

The performance evaluation of single-class (here for category "person") MCN networks are shown in Table 2. Three MCN output visualizations can be found in Figure 6. Networks with human pose estimation are always paired with detection, as detection is required to instantiate humans. Human pose detection is evaluated under the COCO mAP metric. Again, the addition of a segmentation map does not decrease performance of other tasks and even improves the human pose estimation tasks performance by 0,5% when compared to a network without segmentation. In contrast to multiclass segmentation and detection (Table 1), the triple task networks' performance does not suffer significantly from the addition of another head.

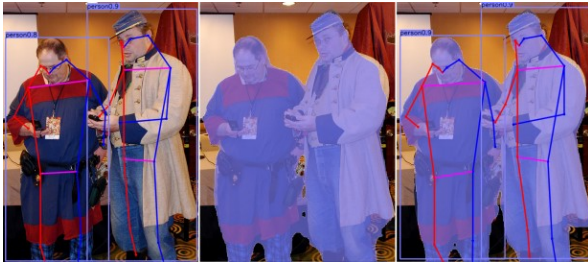

Figure 6: A qualitative analysis of single-class MCN. Human pose detection (left), human segmentation (center) and human pose detection + segmentation (right) networks are visualized on a test sample.

| Network | Segmentation mIoU | Detection mAP |
|---|---|---|
| Only Seg | **49,0%** | NA |
| Only Det | NA | 36.3% |
| Seg + Det | 48,9% | **36.7%** |

Table 1: Single Task (Seg and Det alone) and Multitasks (Seg + Det) MCNs trained and inferenced for all classes of COCO. A multitask network can reach similar performance values as a single task network for either only detection or segmentation.

| Network | Segmentation IoU | Detection AP | Pose mAP |
|---|---|---|---|
| Only Seg | **74,4%** | NA | NA |
| Only Det | NA | 45,0% | NA |
| Seg + Det | 72,8% | **45,2%** | NA |
| Det + Pose | NA | 42,8% | 53,6% |
| Seg + Det + Pose | 74,3% | 42,0% | **54,1%** |

Table 2: Single Task (Seg and Det alone) and Multiple Tasks (Seg + Det, Det + Pose and Seg + Det + Pose) MCNs trained and inferenced only for single (human) class of COCO. For the human class, segmentation and detection heads show similar performance values with and without other heads.

| Network | Segmentation IoU | Detection AP |
|---|---|---|
| Only Seg (multi-class - single-class) | 72,3% - **74,4%** | NA |
| Only Det (multi-class - single-class) | NA | 46,1% - 45,0% |
| Seg + Det (multi-class - single-class) | 72,7% - 72,8 % | **46,3%** - 45,2% |

Table 3: Single Task (Seg and Det alone) and Multiple Tasks (Seg + Det) MCNs trained and inferenced for all classes (multi-class) and single human class (single-class) of COCO for comparison.

Like in multiclass networks, segmentation and detection performs similar alone as together. When pairing detection and human pose estimation, the detection performance falls. The performance of human pose estimation benefits slightly from the addition of a segmentation head.

Third, we compare segmentation and detection performances for the category 'person' only. Network results are given for multiclass and single-class heads (in brackets) in comparison in a single line. When training more than a single class, the performance for segmentation slightly decreases while detection slightly surges. However, the performance difference is rather small. Meanwhile, 79 additional classes are learned with the same backbone.

To test a mixed model, a network with multiclass detection, multiclass segmentation and single-class human pose estimation is also trained. With 47,0% mIoU and 30,2% mAP the network doesn't reach the performance values of multiclass detection/segmentation without human pose annotation in a similar manner as it happens in the single-class triple head network. It also falls behind the single-class triple multitask network in human pose mAP (14,2%). The class imbalance of the setup decreases this metric significantly, as only 66k images contain pose annotations. The application of gradients to non-human classes in detection and segmentation may make training of human pose estimation difficult. All discussed network setups can be seen in a qualitative comparison on test sample in Appendix A1-A2.

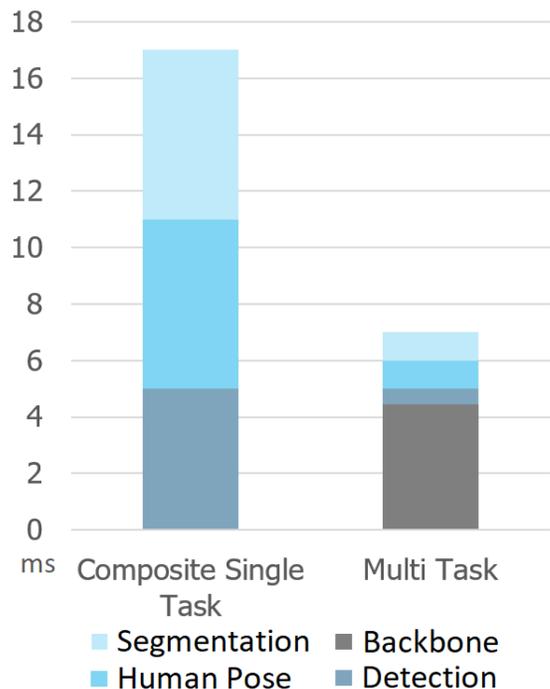

Figure 7: Inference time with ResNet18. ResNet18 is a small, comparably lightweight backbone. A forward pass takes 17ms (58fps) with composite single task network (STN). In contrast, a triple multitask MCN achieves a fast inference time of 7ms (142fps).

Finally, inference times are listed for the network dissemination. Networks are trained with a ResNet18 backbone and evaluated on a Nvidia Tesla V100 GPU. Due to the network heads each only consuming a fraction of the time, different MCN architectures have similar inference times. For the semantic segmentation head, a resolution of 128x128 is chosen, as high upsampling rates consume relatively high computational resources. When evaluating this speed-optimized network, a GPU processing of up to 142fps is possible with a 3-task network (Figure 7). For the more performant, but slower DLA-34 backbone inference times can be found in Table 4. This network uses a 512x512 segmentation output. It is measured on a single Nvidia Tesla P100 GPU. By using a multitask architecture, the framerate in fps can be increased more than twofold.

The network size (in million parameters) increases almost threefold when comparing it to a mixed single task architecture (Table 4). Storage space, for example on an edge device, as well as training times are hence improved respectively.

A more thorough evaluation of inference times of MCNs on different hardware platforms can be found in [24].

| Network | ms | fps | params |
|---|---|---|---|
| Only Seg | 24 | 41.6 | 20.37 |
| Only Det | 19 | 52.6 | 20.40 |
| Det + Pose | 22 | 45.4 | 20.44 |
| Seg + Det | 26 | 38.4 | 20.40 |
| Seg + Det + Pose | 29 | **34.4** | **20.52** |
| Only Seg & Only Det (STN) | 43 | 23.2 | 40.77 |
| Only Seg & Only Det & Only Pose (STN) | 61 | **16.3** | **60.81** |

Table 4: Inference times for MCNs with a DLA-34 backbone and high resolution (512x512) semantic segmentation maps. Trainable parameters (in million) are given in the last row. In this accuracy-optimized approach, a three-task MCN is more than twice as fast as three single task networks (STNs) and almost 3 times as small.

## 5. Conclusion

In many applications, which require the differentiation of tasks into several subtasks, multitask networks can solve them by using a mutual backbone. There is a broad field of potential applications of multitask networks in computer vision, such as autonomous driving and medical imaging, for which we propose the Multitask-CenterNet (MCN) architecture. The MCN can dramatically reduce inference times and network size by sharing most of the network layers and reducing the latency and number of parameters. At the same time, performance values can remain on a high level even when several heads make use of a backbone and partially even outperform single task networks. By comparing networks with various heads, we also show which heads have a high influence on performance of other heads and where class relations play a role in performance losses. An imbalanced class set decreases performance, an effect that could be avoided when setting coherent dataset annotation rules. Meanwhile, the addition of heads in general usually does not decrease performance even when most of the parameter count is the same in the backbone.

We therefore conclude that when predicting labels from a similar domain on the same input domain (vision), a multitask network like the MCN is a coherent architectural choice.

## 6. Acknowledgements

This work is in part funded by the German Federal Ministry for Economic Affairs and Energy (BMWi) through the grant 19A19013Q, project KI Delta Learning.

| | | | |
|---|---|---|---|
| Multiclass Segmentation | 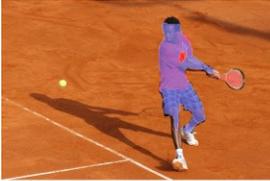 | 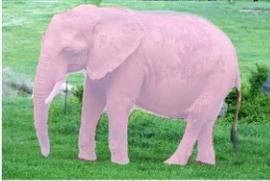 | 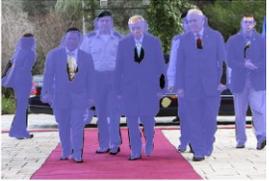 |
| Single-class Segmentation | 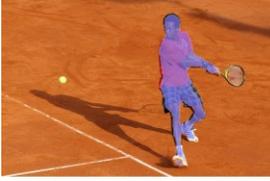 | 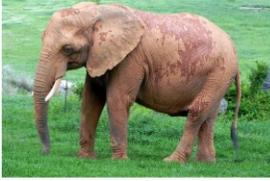 | 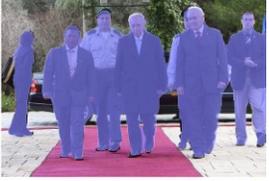 |
| Multiclass Detection | 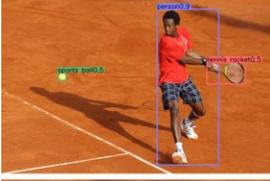 | 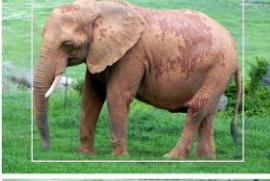 | 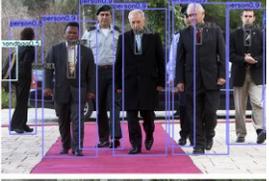 |
| Single-class Detection | 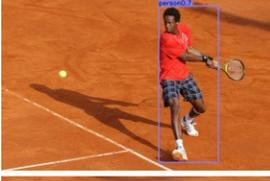 | 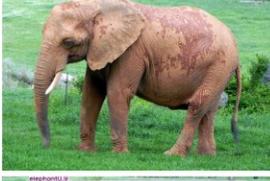 | 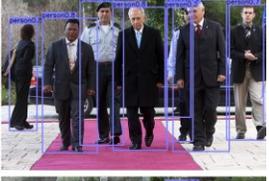 |
| Multiclass Segmentation, Multiclass Detection | 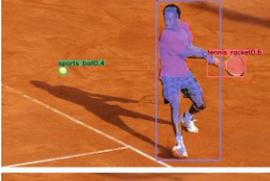 | 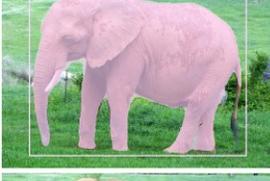 | 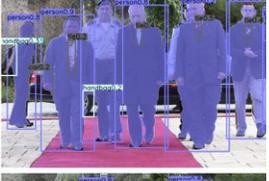 |
| Single-class Segmentation, Single-class Detection | 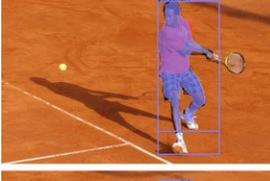 | 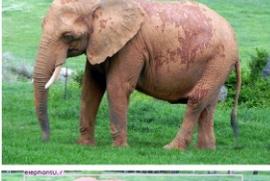 | 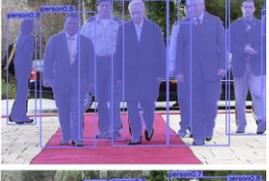 |
| Multiclass Segmentation, Multiclass Detection, Human Pose Estimation | 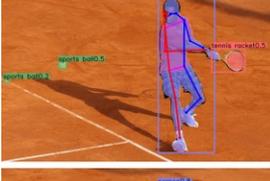 | 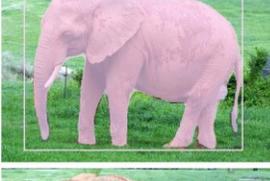 | 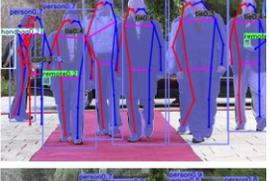 |
| Single-class Segmentation, Single-class Detection, Human Pose Estimation | 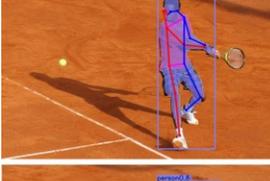 | 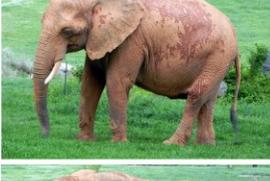 | 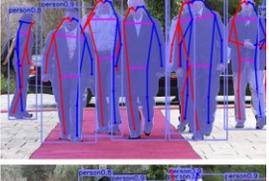 |
| Single-class Detection, Pose Estimation | 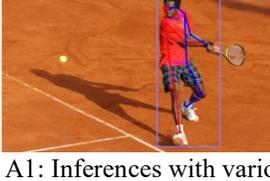 | 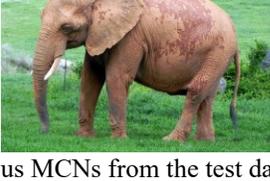 | 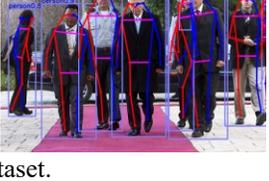 |

Appendix A1: Inferences with various MCNs from the test dataset.

| | | | | |
|---|---|---|---|---|
| Multiclass Segmentation | 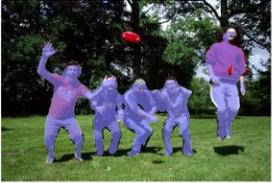 | 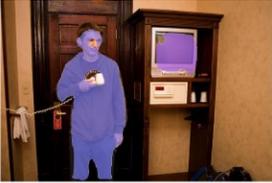 | 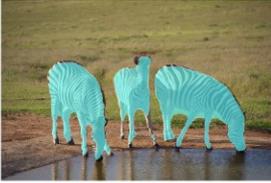 | 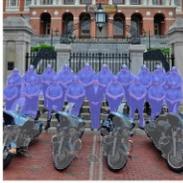 |
| Single-class Segmentation | 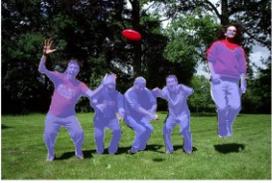 | 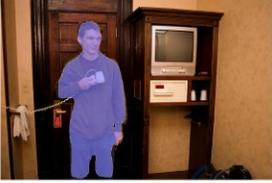 | 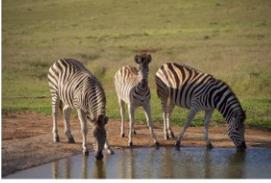 | 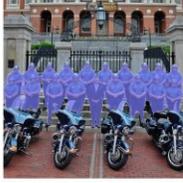 |
| Multiclass Detection | 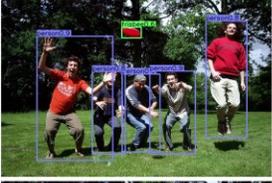 | 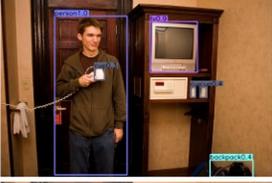 | 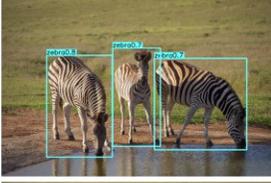 | 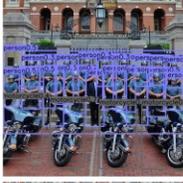 |
| Single-class Detection | 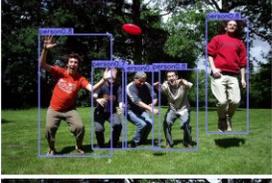 | 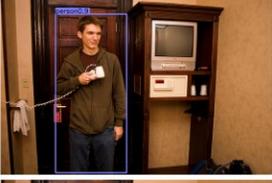 | 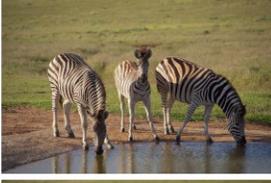 | 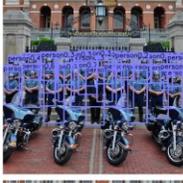 |
| Multiclass Segmentation, Multiclass Detection | 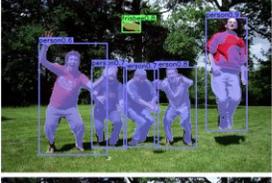 | 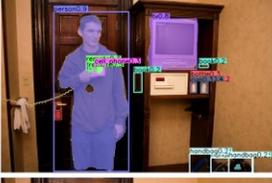 | 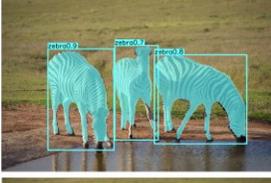 | 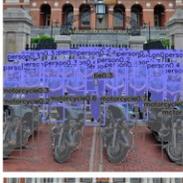 |
| Single-class Segmentation, Single-class Detection | 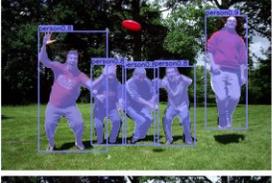 | 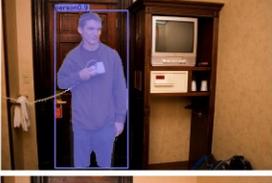 | 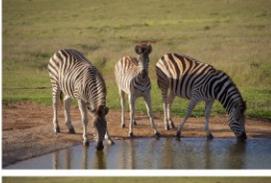 | 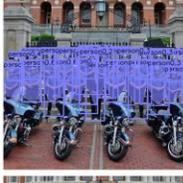 |
| Multiclass Segmentation, Multiclass Detection, Human Pose Estimation | 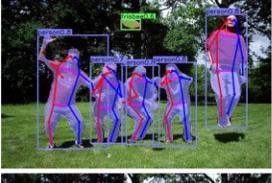 | 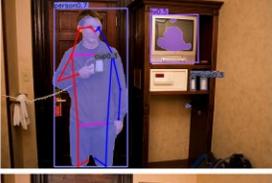 | 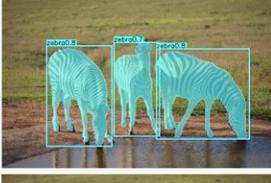 | 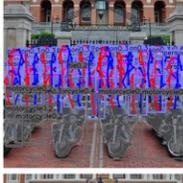 |
| Single-class Segmentation, Single-class Detection, Human Pose Estimation | 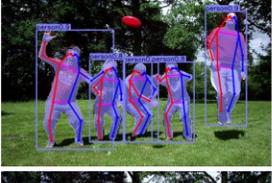 | 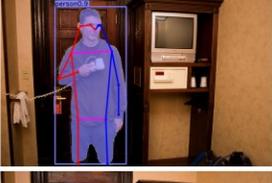 | 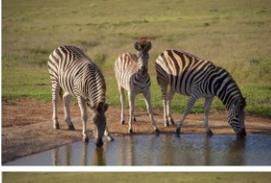 | 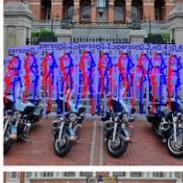 |
| Single-class Detection, Pose Estimation | 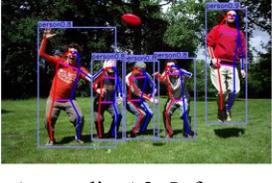 | 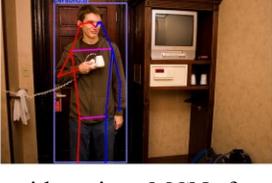 | 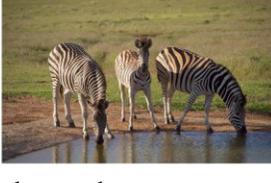 | 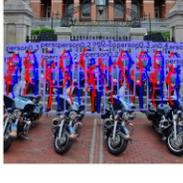 |

Appendix A2: Inferences with various MCNs from the test dataset.